\documentclass[review,3p]{elsarticle}

\usepackage{hyperref}

\usepackage{times}
\usepackage{latexsym}
\usepackage{amsmath,amssymb,bbm}
\usepackage{bm}
\usepackage{booktabs}
\usepackage{graphicx}
\usepackage{multirow}
\usepackage{arydshln}
\usepackage{url}
\usepackage{pgfplots}
\usepackage{ dsfont }
\usepackage{latexsym}
\usepackage{subfigure}
\usepackage{array}
\usepackage[numbers]{natbib}
\definecolor{blue1}{RGB}{5, 112, 216}
\journal{Neurocomputing}

\begin{document}
	
	\begin{frontmatter}
		
		\title{Dialogue-adaptive Language Model \\Pre-training From Quality Estimation}

		\author[mymainaddress,anotheraddress]{Junlong Li}
		\ead{lockonn@sjtu.edu.cn}
		
		\author[mymainaddress,anotheraddress]{Zhuosheng Zhang}
		\ead{zhangzs@sjtu.edu.cn}
		
		\author[mymainaddress,anotheraddress]{Hai Zhao\corref{mycorrespondingauthor}}
		\cortext[mycorrespondingauthor]{Corresponding author. This paper was partially supported by Key Projects of National Natural Science Foundation of China (U1836222 and 61733011).}
		\ead{zhaohai@cs.sjtu.edu.cn}
		
		\address[mymainaddress]{Department of Computer Science and Engineering, Shanghai Jiao Tong University}
		\address[anotheraddress]{Key Laboratory of Shanghai Education Commission for Intelligent Interaction \\and Cognitive Engineering, Shanghai Jiao Tong University}

		\begin{abstract}
			Pre-trained language models (PrLMs) have achieved great success on a wide range of natural language processing tasks by virtue of the universal language representation ability obtained by self-supervised learning on a large corpus. These models are pre-trained on standard plain texts with general language model (LM) training objectives, which would be insufficient to model dialogue-exclusive attributes like \textit{specificity} and \textcolor{black}{\textit{informativeness}} reflected in these tasks that are not explicitly captured by the pre-trained universal language representations.
			In this work, we propose dialogue-adaptive pre-training objectives (DAPO) derived from quality estimation to simulate dialogue-specific features, namely coherence, specificity, and informativeness. As the foundation for model pre-training, we synthesize a new dialogue corpus and build our training set with two unsupervised methods: 1) \textit{coherence}-oriented context corruption, including utterance ordering, insertion, and replacement, to help the model capture the coherence inside the dialogue contexts; and 2) \textit{specificity}-oriented automatic rescoring, which encourages the model to measure the quality of the synthesized data for dialogue-adaptive pre-training by considering \textit{specificity} and \textit{informativeness}. 
			Experimental results on widely used open-domain response selection and quality estimation benchmarks show that DAPO significantly improves the baseline models and achieves state-of-the-art performance on the MuTual leaderboard, verifying the effectiveness of estimating quality evaluation factors into pre-training.
		\end{abstract}
		
		\begin{keyword}
			Pre-trained Language Models\sep Dialogue-adaptive Pre-training\sep Dialogue Quality Estimation\sep Open-domain Dialogue Systems.
		\end{keyword}
		
	\end{frontmatter}
	
	
	\section{Introduction}
	Pre-trained language models (PrLMs) have achieved impressive performance in a series of natural language processing tasks. Some prominent examples of PrLMs are BERT \cite{devlin2019bert}, GPT \cite{radford2018improving}, XLNet \cite{yang2019xlnet}, RoBERTa \cite{liu2019roberta}, ERNIE \cite{ernie2019sun, ernie202019sun}, ALBERT \cite{lan2019albert} and ELECTRA \cite{electra2020Clark}. \textcolor{black}{They can also be used in more expansive fields like Bioinformatics \cite{le2021deep,le2021transformer}.}
	The PrLMs are commonly employed through a \textit{pre-training then fine-tuning} paradigm: the models are first trained on large-scale unlabeled task-independent corpora with general training objectives, like masked language modeling (MLM) \cite{ClozePA1953Taylor} or next sentence prediction (NSP) \cite{devlin2019bert}, to learn universal language representations; then, the trained models are fine-tuned on the smaller datasets of downstream tasks by leveraging extra task-specific modules for adaption.
	However, the learned universal language representations for dialogue tasks would not sufficiently and accurately cover the dialogue-aware features because the literary style of dialogues and plain texts varies dramatically. More specifically, as opposed to general plain texts, dialogue involves multiple speakers, intentions, topics; thus, the utterances are full of transitions. As a result, directly fine-tuning PrLMs on the dialogue datasets would be sub-optimal to model dialogues that contain exclusive attributes.
	
	To help PrLMs adapt to dialogue-related tasks, recent studies have investigated further dialogue-adaptive pre-training on PrLMs before fine-tuning them on dialogue tasks. \cite{DialoGPTLG2020Zhang} pre-trained GPT further with a conventional Language Model (LM) objective on a large dialogue corpus, Reddit, and get DialoGPT for response generation tasks; \cite{zhang2021kkt} and \cite{AnED2019Whang} pre-trained BERT with Mask Language Model (MLM) objective on the target datasets before fine-tuning on response selection tasks. Besides merely pre-training on dialogue datasets with general objectives, some 
	studies propose particular auxiliary tasks or objectives for dialogue-adaptive pre-training. Notably, \cite{wu2020tod} proposed a response contrastive loss to match the context with the corresponding response and distinguish from the randomly sampled negative response. \cite{xu2020learn_aux} employed auxiliary objectives to measure the utterance consistency of a dialogue session and restore the corrupted utterances when fine-tuning models for response selection tasks.

	Despite the progress made by the methods mentioned above, the guideline of dialogue-adaptive pre-training is still not fully exploited, especially for open-domain dialogues as the main focus of this work. The major focus of existing dialogue-adaptive pre-training strategies revolves around merely one attribute, \textit{coherence}, either on token-level or sentence-level. Therefore, the previous studies can be referred to as \textit{coherence}-oriented objectives \cite{xu2020learn_aux}, which are commonly implemented in a discriminative way: corrupting the dialogue context by masking, randomly sampling, or re-ordering as \textit{negative} samples, and restoring the original context as the \textit{positive} ones. 
	
	
	
	\begin{figure}
		\centering
		\includegraphics[scale=0.36]{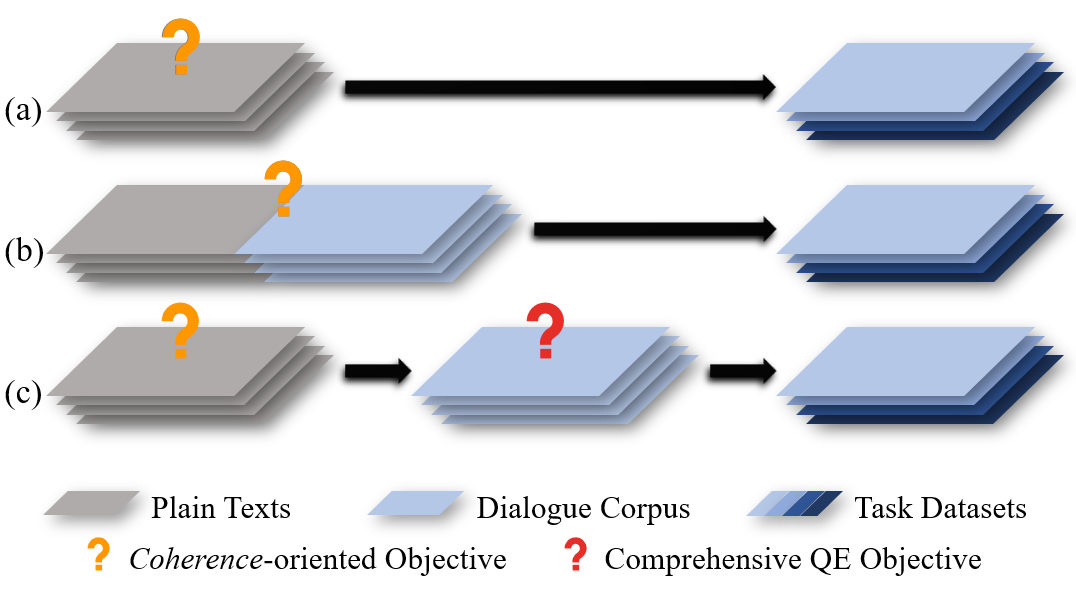}
		\caption{\textbf{Comparison between two previous workflows and ours:} (a) Original PrLM workflow. (b) Existing dialogue-adaptive PrLM workflow. (c) Our dialogue-adaptive PrLM workflow, where QE stands for Quality Estimation, and we instantiate the QE objective as DAPO in this work.}
		\label{fig:overll_idea}
	\end{figure}
	
	As opposed to written language texts, dialogues, as spoken language texts, are full of redundant or uninformative utterances. As it is pointed out by recent dialogue evaluation studies \cite{WhatMA2019See,UnsupervisedEO2020Mehri,mehri2020usr,pang2020towards,DesigningPA2020Zhao}, estimating whether an open-domain dialogue is \textit{informative} is also essential for assessing its quality, which motivates us to improve dialogue-adaptive pre-training by considering informativeness to better simulate dialogue-specific characteristics. In addition, measuring the informativeness of the utterance would provide a more fine-grained self-supervision like informativeness degree during training, and guide the model to get rid of meaningless utterances. Therefore, in this work, we make a first attempt to bridge comprehensive dialogue quality estimation and pre-training and present dialogue-adaptive pre-training objectives (DAPO) by considering the salient characteristics, including coherence and specificity. The overall workflow of utilizing DAPO and the comparison between existing methods are shown in Figure \ref{fig:overll_idea}.  A detailed explanation of the terms are shown as follows:
	\begin{enumerate}
		\item \textit{coherence}: whether a dialogue is coherent in its semantics and logic so that readers feel easy to read it.
		\item \textit{specificity}: whether the tokens, phrases, and expressions in a dialogue are specific and diverse to avoid being dull and monotonous.
	\end{enumerate}
	
	As the foundation for model pre-training, we first synthesize a new open-domain dialogue corpus used for pre-training with each dialogue in it as a positive sample to facilitate discriminative pre-training. Inspired by \cite{ModelingLC2005Barzilay,CoherenceMF2018Cervone,DialogueCA2020Mesgar}, we propose \textit{coherence}-oriented context corruption, including utterance ordering, utterance insertion and utterance replacement, to generate incoherent dialogues as negative samples.\footnote{The negative samples are scored as 0 while the positive ones as 1.} Next, we do \textit{specificity}-oriented automatic rescoring by multiplying the scores of positive samples with a \textit{token-specificity} coefficient measured by \textit{n}-gram normalized inverse document frequency (\textit{N}-NIDF) to further distinguish the samples by quantifying how specific they are. Finally, the PrLMs are trained on all these unsupervised annotated samples with a regression task.\footnote{\textcolor{black}{Our codes are publicly available in  \url{https://github.com/lockon-n/DAPO}}.}

	In summary, our contributions in the paper are three-fold:
	\begin{enumerate}
		\item We bridge the gap between dialogue-adaptive pre-training and dialogue estimation to facilitate a new research line of dialogue-adaptive pre-training from quality estimation by considering the salient characteristics, including coherence and informativeness.
		\item To simulate dialogue-specific features, we propose dialogue-adaptive pre-training objectives
		including coherence-oriented context corruption and token-specificity rescoring, for open-domain dialogue modeling; 
		\item We empirically verify that the quality estimation methods improve PrLMs on both response selection and quality evaluation tasks by modeling the diverse aspects of dialogue-related characteristics.
	\end{enumerate}

	\section{Background and Related Works}

	\subsection{Pre-trained Language Models}
	Recent works have explored various architecture
	choices and training objectives for large-scale LM
	pre-training. Most of the PrLMs are based on the encoder in Transformer, among which Bidirectional
	Encoder Representations from Transformers
	(BERT) \cite{devlin2019bert} is one of the most
	representative works. It adopts masked language modeling (MLM) and next sentence prediction (NSP) as its pre-training objectives. MLM is also referred as a \textit{Cloze} task. It first masks out some tokens from the input sentences and then trains the model to predict them by the rest of the tokens. NSP is another widely used pre-training objective. It trains the model to distinguish whether two input sentences are continuous segments from the training corpus. Several subsequent variants have been proposed further to enhance the capacity of PrLMs, such as XLNet \cite{yang2019xlnet} trained with a permutation language model objective, ALBERT \cite{lan2019albert} trained with MLM and sentence order prediction, and ELECTRA \cite{electra2020Clark} trained with replaced token detection to distinguish between original and generated tokens. 
	

	\subsection{Dialogue-adaptive Pre-training}
	To make PrLMs more compatible in the dialogue scenario and corresponding downstream task, existing works have tried to do adaptive further pre-training on another dialogue corpus or the target task dataset. A part of the studies perform training on large conversational
	data like Reddit for response selection or generation tasks \cite{wolf2019transfer,bao2020plato,henderson2020convert,DialoGPTLG2020Zhang,zhang2021kkt,gu2021deep,AnED2019Whang,su2021prototype},
	with conventional objectives such as MLM and NSP. There are also works that take a step further on the basis of traditional losses. \cite{Kumar2020dar} and \cite{gu2020dialogbert} predict the token order within utterances as well as the utterance order in the dialogues. TOD-BERT \cite{wu2020tod} trains BERT further with the combination of a newly-designed Response Contrastive loss and MLM for task-oriented dialogues. \cite{xu2020learn_aux} do multi-task joint learning with four self-supervised tasks: session-level matching, utterance restoration, incoherence detection, and consistency classification as auxiliary objectives when fine-tuning models for response selection tasks.
	
	Although with various forms of pre-training objectives, all these models, no matter general or dialogue-adaptive ones, only concern the \textit{coherence} aspect for modeling dialogues. The main difference between them is just the numbers and types of granularities. For open-domain dialogues emphasized in this work, many more characteristics should be taken into account as revealed by recent dialogue evaluation works \cite{WhatMA2019See,UnsupervisedEO2020Mehri,mehri2020usr,pang2020towards,DesigningPA2020Zhao}, so we are motivated to model \textit{specificity} together with \textit{coherence} as a way of fusing quality estimation into pre-training. To avoid redundant comparison with all the variants of the above mentioned \textit{coherence}-oriented objectives, we pick the most representative ones, MLM and NSP (one for token-level and the other for sentence-level), in the experimental part to show our advantages over them.
	
	\section{Dialogue Adaption Pre-training Objective from Quality Estimation}
	\subsection{Pre-training Corpus Construction}
	Existing large dialogue corpora such as Reddit \cite{DialoGPTLG2020Zhang} or Ubuntu \cite{lowe2015ubuntu} are directly crawled from the Internet forums without further processing. A considerable proportion of sentences in these large corpora do not follow grammatical standards or even have syntactic errors and spelling mistakes. As a result, we avoid using them and choose to construct a new open-domain dialogue corpus based on four manually-proofread, medium-size datasets: DailyDialog \cite{DailyDialogAM2017Li}, PERSONA-CHAT \cite{PersonalizingDA2018Zhang}, Topical-Chat \cite{TopicalChatTK2019Gopalakrishnan}, and BlendedSkillTalk \cite{CanYP2020Smith}. The total number of dialogues from these datasets is 49,930. Dialogues extracted from these datasets with more than 10 utterances are split into several consecutive, overlapping dialogue segments to prevent the length of text from exceeding the \texttt{max-sequence-length} of models, while the others stay intact. As a result, all dialogues in our corpus have no more than 10 utterances, and they are regarded as positive samples.
	
	\subsection{\textit{Coherence}-oriented Context Corruption}
	We generate incoherent negative dialogue samples through \textit{coherence}-oriented context corruption, including:  utterance ordering (UO) and utterance insertion (UI) and utterance replacement (UR) \cite{ModelingLC2005Barzilay,CoherenceMF2018Cervone,DialogueCA2020Mesgar}. 
	
	\begin{enumerate}
		\item \textbf{UO}: The order of utterances in a dialogue is permuted randomly.
		\item \textbf{UI}: One utterance in a dialogue is removed and then re-inserted in any possible position except the original one in the dialogue.
		\item \textbf{UR}: One of the utterances in a dialogue is replaced with another utterance that is randomly selected from other dialogues.
	\end{enumerate}
	
	For each positive sample, we construct three negative samples by \textit{coherence}-oriented context corruption, i.e., one for each operation. After that, we score all the negative samples as 0 and positive ones as 1 to distinguish whether a sample is coherent or not.
	
	The corpus is further split into train and dev sets with a ratio of 0.9/0.1. More detailed statistics of our corpus can be found in Table 1. 
	
	\begin{table}[h]
		\renewcommand\arraystretch{0.8}
		\footnotesize
		\centering
		{
			\begin{tabular}{lrr}
				\toprule
				& \textbf{Train} & \textbf{Dev}\\
				\midrule
				\# of all samples                 & 1045K       &116K     \\
				\# of positive samples            & 261K        &29K     \\
				\# of negative samples            & 784K        &87K     \\
				avg. \# utter. per sample         & 9.84        &9.84     \\
				avg. \# tokens per sample         & 177.09      &177.30     \\      
				\bottomrule       
			\end{tabular}
		}
		\label{tb:data-stat}
		\caption{\textbf{Data Statistics}: Our open-domain \\dialogue dataset for pretraining.}
	\end{table}
	
	\subsection{\textit{Specificity}-oriented Automatic Rescoring}\label{sec:score_nidf}
	To take a comprehensive quality estimation factor into the pre-training objective, we further leverage \textit{specificity}-oriented automatic rescoring. The scores of positive samples are additionally multiplied by a \textit{token-specificity} coefficient to judge how much they are \textit{specific} and \textit{informative}. This coefficient is measured by \textit{n}-gram normalized inverse document frequency (\textit{N}-NIDF), which is extented from normalized inverse document frequency (NIDF) \cite{WhatMA2019See} and it has been shown effective for reflecting token rareness. The inverse document frequency (IDF) of an \textit{n}-gram $ng$ is
	\begin{equation}
		\textrm{IDF}(ng)=\textrm{log}(D/{c_{ng}^D}),
	\end{equation}
	where $D$ is the number of the original dialogues from the four dialogue datasets (i.e. $D$ = 49,930), and ${c_{ng}^D}$ is the number of those dialogues that contain $ng$. Then, normalized IDF (NIDF) for $ng$ is as follows:  
	\begin{equation}
		\begin{aligned}
			&\textrm{NIDF}(ng)=
			\frac{\textrm{IDF}(ng)-\textbf{\textrm{min-idf}}}{\textbf{\textrm{max-idf}}-\textbf{\textrm{min-idf}}},
		\end{aligned}
	\end{equation}
	where \textbf{min-idf} and \textbf{max-idf} are the minimum and maximum of all IDFs. The \textit{N}-NIDF of a sample $s$ is the weighted mean for all NIDF of \textit{n}-grams in this sample:
	\begin{equation}
		\begin{aligned}
			&N\textrm{-NIDF}(s)=\sum_{ng\in s_{ng}}\textrm{NIDF}(ng)\times\frac{c_{ng}^s}{|s_{ng}|},
		\end{aligned}
	\end{equation}
	where $s_{ng}$ denotes all the \textit{n}-grams in this sample, and $c_{ng}^s$ denotes the times $ng$ appears in $s$.  
	
	For each positive sample, we calculate \textit{N}-NIDF of it with \textit{n} = 3, and use it as the \textit{token-specificity} coefficient. Consequently, the final scores for all the samples in our corpus is as follows:
	$$ \textrm{score}(s)=\left\{
	\begin{aligned}
		&0                             &\ &s\in S^- \\
		&1\times\textrm{3-NIDF}(s)\in[0,1]  &\ &s\in S^+ \\
	\end{aligned}
	\right.
	$$ where $S^-$ and $S^+$ denote the negative and positive samples respectively.
	
	This unsupervised annotation is able to measure \textit{coherence} , \textit{specificity} and \textit{informative} simultaneously as an overall quality estimation, while it does not require any time- and cost-intensive human labeling, allowing us to take full advantage of the large-scale unlabeled corpus.
	
	\begin{figure}
		\centering
		\includegraphics[scale=0.4]{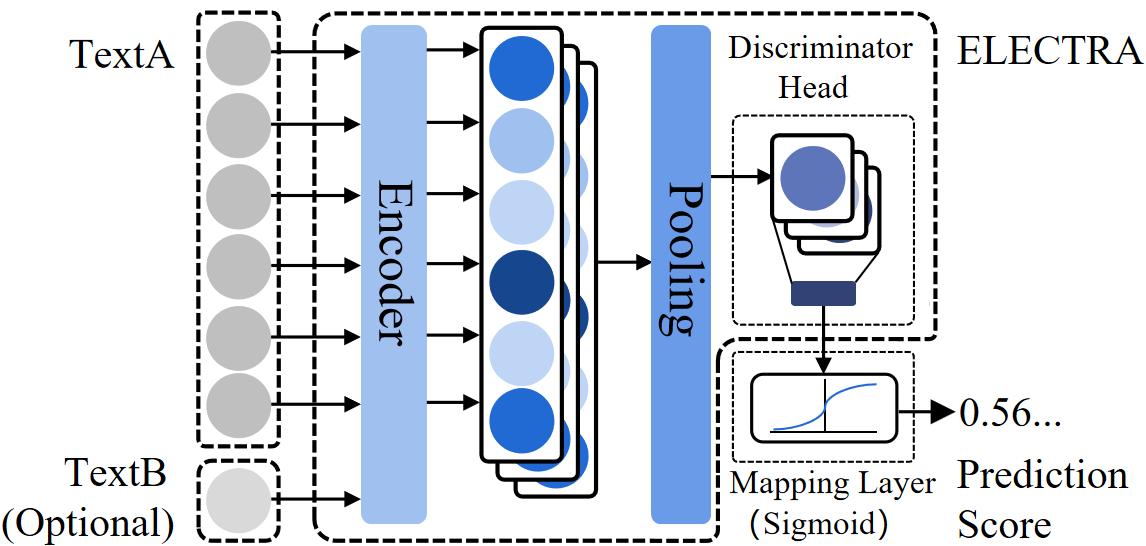}
		\caption{\textbf{An overview of our model}: TextA is the dialogue history, and TextB is the last utterance or the candidate response. Contents in the large dotted box are the ELECTRA model, and its outputs are sent into the Sigmoid mapping layer to get a score in $[0,1]$. }
		\label{fig:overll_structure}
	\end{figure}
	
	\subsection{Model Implementation}
	The discriminator of $\textrm{ELECTRA}_{large}$ \cite{electra2020Clark} is the PrLM adopted in our work and referred to as ELECTRA for brief in the following statements. \textcolor{black}{We drop the generator part of it so that it has the same structure as BERT with an additional discriminator head. ELECTRA requires a \textit{textA} and an optional \textit{textB} as the inputs, and insert an [SEP] token between them if \textit{textB} exists.
		By setting the number of category as 1, the discriminator head can map the representation of the [CLS] token to a real value.} For DAPO, we regard each of our samples as a long text sequence and input it into ELECTRA as \textit{textA} while leaving \textit{textB} as blank. To match the range of scores, a mapping layer is added on top of ELECTRA. It consists of a sigmoid function, transforming the original output logit of PrLMs to a real number ranging from 0 to 1. The overall structure is shown in Figure \ref{fig:overll_structure}. 
	
	During pre-training on our open-domain dialogue corpus, the parameters are updated by mean-square error (MSE) loss:
	\begin{equation}
		\begin{aligned}
			\mathbb{L}_{\rm MSE}=\frac{1}{b}\sum_{i=1}^b{(s_i-\hat{s_i})^2},
		\end{aligned}
	\end{equation}
	where $b$ is the batch size, $s_i$ and $\hat{s_i}$ denote the real score and the prediction score of a sample, respectively.
	
	\section{Settings}
	\subsection{\textcolor{black}{Task Description and Datasets}}
	We evaluate our method on two typical kinds of open-domain dialogue tasks:  Response Selection (RS) and Quality Evaluation (QE).


	
	\paragraph{Response Selection (RS)} This task requires models to select the best response from some candidates with a given dialogue history. Recall at position \textit{n} in candidates (R@n) and Mean Reciprocal Rank (MRR) \cite{TheTQ1999Voorhees} are set to be the evaluation metrics. 
	
	MuTual \cite{MuTualAD2020Cui} is selected as the dataset. It consists of 8,860 manually annotated open-domain dialogues based on Chinese student English listening comprehension exams and requires models to handle various reasoning problems. Experiments are also conducted on the advanced version of it, $\textrm{MuTual}^{plus}$, where one of the candidate responses is replaced by a safe response (e.g., \textit{Could you repeat that?} or \textit{I'm really sorry, I didn't catch that.}) for each example. If the original right answer is replaced, then the safe response becomes the best one; otherwise, the original positive response is still the best one. The introduction of safe responses makes MuTual$^{plus}$ more challenging than MuTual.
	
	
	\paragraph{Quality Evaluation (QE)} 
	This task needs the model to evaluate a specific response after a given dialogue history by assigning a score for this reponse. Each example in the task datasets has been labeled with a set of human judgment scores by several annotators to determine the overall impression of this response \cite{mehri2020usr,pang2020towards,DesigningPA2020Zhao,WhatMA2019See,UnsupervisedEO2020Mehri,DBLP:journals/corr/abs-2106-03706}. Following previous studies, we use Pearson and Spearman correlation to examine whether the prediction scores are correlated with human judgments.
	
	A subset of dialogues from DailyDialog \cite{DailyDialogAM2017Li} and PERSONA-CHAT \cite{PersonalizingDA2018Zhang} respectively annotated in a previous study \cite{DesigningPA2020Zhao} are chosen for response-level evaluation datasets. These two datasets provide the human judgments of the \textit{Overall Quality} for each example. We also try directly applying our model without further training as an individual dialogue evaluation module and choose a subset of the repository mentioned in \cite{DBLP:journals/corr/abs-2106-03706} as the datasets. Since human judgment scores may have a range unmatched with 0-1 (e.g., 1-5 or 0-3), they are uniformly mapped into the range of 0-1 to match the prediction scores.
	
	\subsection{Task-specific Fine-tuning}
\begin{figure}
		\centering
		\includegraphics[scale=0.55]{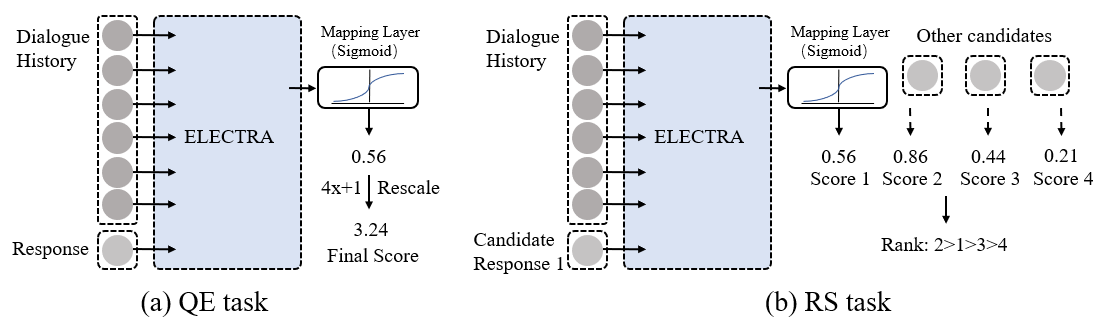}
		\caption{\textcolor{black}{\textbf{An overview of the fine-tuning adaption}: The model architecture is the same as the pre-training stage, except that we need to process all the candidate responses in RS tasks.}}
		\label{fig:ft}
	\end{figure}

	The model architecture for fine-tuning is almost the same as the one for DAPO pre-training shown in Figure \ref{fig:overll_structure}. \textcolor{black}{Here, we show how to adapt it to RS and QE tasks. An overview of the adaption for fine-tuning is shown in Figure \ref{fig:ft}.}
	
	\paragraph{Response Selection (RS)} The dialogue history is input as \textit{textA} and each candidate response is input as \textit{textB} with an [SEP] token in the middle.
	For responses in a dialogue example, the model will output corresponding scores, and use these values to rank them from large to small. \textcolor{black}{The loss function is still MSE as we label the correct response as 1 and others as 0.}
	
	\paragraph{Quality Evaluation (QE)} \textcolor{black}{For QE tasks, the inputs are the same as the RS task. Still, the prediction score with the mapping layer for each example is used, and we rescale the score to the original range for evaluation, for example to a range of [1,5]. }

	
	

	\subsection{Baseline Models}\label{sec:baseline}
	Pre-trained ELECTRA without any further task-specific pre-training is used as one of our baselines. To show the effectiveness of our proposed DAPO, we follow the same steps of our method by replacing DAPO with MLM and NSP to get ELECTRA-MLM and ELECTRA-NSP as the strong and representative baselines of \textit{coherence}-oriented objectives. The pre-training data of MLM is directly the original corpus, and for each sentence pair in the corpus, we generate a negative sample by replacing the second utterance with a random one to get the pre-training data for NSP. The corpora for these two objectives are also split into train/dev set with the ratio of 0.9/0.1. We also combine MLM and NSP together (i.e., ELECTRA-MLM+NSP) like BERT and use it as a baseline model. 
	
	Besides our implementation, we compare with the following public works. Some of the results are from corresponding leaderboards.
	
	\begin{table}
		\renewcommand\arraystretch{0.8}
		\centering
		\footnotesize
		\setlength{\tabcolsep}{5.5pt}
		\label{tb:mutual}
		{
			\begin{tabular}{lllllll}
				\toprule
				
				\textbf{Model}
				&\multicolumn{3}{c}{\textbf{MuTual}} &\multicolumn{3}{c}{\textbf{MuTual$^{plus}$}} 
				\\
				
				&\textbf{R@1} & \textbf{R@2} & \textbf{MRR} & \textbf{R@1} & \textbf{R@2} & \textbf{MRR} \\
				\midrule
				\textit{On Leaderboard} \\
				RoBERTa-MC &0.686	&0.887	&0.822	&0.643	&0.845	&0.792
				\\
				RoBERTa &0.713	&0.892	&0.836	&0.626	&0.866	&0.787\\
				RoBERTa+OCN &0.867	&0.958	&0.926	&-	&-	&- \\
				GRN-v2 &0.915	&0.983	&0.954	&\textbf{0.841}	&\textbf{0.957}	&\textbf{0.913}\\
				MDFN &\textbf{0.916}	&\textbf{0.984} &\textbf{0.956}	&-	&-	&-
				\\
				\midrule
				\textit{In Paper} \cite{MuTualAD2020Cui}& & & & & \\
				Dual LSTM      & 0.266 & 0.528 & 0.538 & 0.266 & 0.528 & 0.538
				\\
				
				SMN            & 0.274 & 0.524 & 0.575 & 0.274 & 0.524 & 0.575
				\\
				
				DAM            & 0.239 & 0.463 & 0.575 & 0.239 & 0.463 & 0.575
				\\
				
				BERT           & 0.657 & 0.867 & 0.803 & 0.657 & 0.867 & 0.803
				\\
				
				BERT-MC        & 0.661 & 0.871 & 0.806 & 0.661 & 0.871 & 0.806

				\\      
				\midrule
				\textit{Our Implementation}                  \\
				ELECTRA        & 0.900 & 0.979 & 0.946 & 0.823 & 0.947 & 0.901
				
				\\
				
				ELECTRA-DAPO    & \textbf{0.916} (+1.6\%) & \textbf{0.988} (+0.9\%)& \textbf{0.956} (+1.0\%)& \textbf{0.836} (+1.3\%)& \textbf{0.955} (+0.8\%)& \textbf{0.910} (+0.9\%)
				
				\\
				
				
				\bottomrule       
			\end{tabular}%
			\caption{\textbf{Main Results on MuTual and MuTual$^{plus}$}: Top 2 scores for each metric are in bold. We do not include all the results on the leaderboard, as some of them are not publicly available now.}
		}	
	\end{table}
	
	
	\paragraph{Response Selection (RS)}
	Individual scoring methods: Dual LSTM \cite{lowe2015ubuntu}, SMN\cite{wu2017sequential}, DAM \cite{zhou2018multi}, BERT \cite{devlin2019bert} and RoBERTa \cite{liu2019roberta}. These models score each response in an example individually. Multi-choice method: including BERT-MC \cite{devlin2019bert}, RoBERTa-MC \cite{liu2019roberta}, OCN \cite{ran2019ocn}, GRN-v2 \cite{liu2020grn} and MDFN \cite{liu2020mdfn}, are multi-choice models that handle all the responses at the same time.
	
	\paragraph{Quality Evaluation (QE)}
	Reference-based metrics: BLEU \cite{BleuAM2002Papineni} (we use the best result among BLEU-1,2,3 and 4), ROUGE \cite{ROUGEAP2004Lin} , METEOR \cite{METEORAA2005Banerjee}, BERTScore \cite{BERTScoreET2020Zhang}, ADEM \cite{TowardsAA2017Lowe}, RUBER \cite{RUBERAU2018Tao}. \textcolor{black}{These methods are commonly-used metrics in sequence-to-sequence tasks by calculating the word overlapping in the generated texts and the reference texts. Following previous works in quality evaluation tasks, we use these metrics to meausre the similarity score between the candidate response and the reference response and use the similarity as the final score.} Reference-free metric: RoBERTa-eval \cite{DesigningPA2020Zhao}, which relies on the powerful PrLM RoBERTa. \textcolor{black}{It utilizes RoBERTa$_{large}$ as its backbone model, which has almost the same number of parameters as the ELECTRA$_{large}$ we use. RoBERTa-eval is first trained in an unsupervised way with an NSP task, and then do supervised training on the human-annotated dialog evaluation dataset.} It evaluates dialogues with no reference responses and directly maps the response representation to a score.

	\subsection{Implementation Details}
	Our code is written based on \texttt{Transformers}\footnote{\url{https://github.com/huggingface/transformers}.} , an open-source github repository. Some baselines in QE tasks are from \cite{DesigningPA2020Zhao}, \cite{BERTScoreET2020Zhang}, and \cite{DBLP:journals/corr/SharmaASZ17}. \textcolor{black}{We use the official tokenizer for ELECTRA in \texttt{Transformers} to tokenize the input text, which is based on punctuation splitting and wordpieces.}
	
	
	We use Adam \cite{AdamAM2015Kingma} as the optimizer with $\epsilon$ = 1e-8 and no weight decay. The learning rate of our task-specific pre-training (DAPO, MLM, and NSP) is 1e-5, batch size per GPU is 10, warmup rate is 0.1, and the max sequence length is 512. We train 5 epochs on our dialogue corpus to get the pre-trained models. Then they are fine-tuned for 8 epochs on the open-domain dialogue tasks with learning rate, batch size, and warmup rate the same as pre-training. We empirically set these hyperparameters without much tuning to show the robustness of our method.

	\begin{table}
		\renewcommand\arraystretch{0.8}
		\centering
		\footnotesize
		\setlength{\tabcolsep}{2pt}
		\label{tb:dd}
		{
			\begin{tabular}{lllllllll}
				\toprule
				
				\textbf{Model} 	&\multicolumn{4}{c}{\textbf{DailyDialog}} &\multicolumn{4}{c}{\textbf{PERSONA-CHAT}}\\
				
				&\multicolumn{2}{c}{\textbf{Dev}} &\multicolumn{2}{c}{\textbf{Test}}
				&\multicolumn{2}{c}{\textbf{Dev}} &\multicolumn{2}{c}{\textbf{Test}}\\
				
				&\textbf{Pearson} &\textbf{Spearman} &\textbf{Pearson} &\textbf{Spearman}&\textbf{Pearson} &\textbf{Spearman} &\textbf{Pearson} &\textbf{Spearman}\\
				\midrule
				\textit{Our Re-running}          \\
				BLEU        & 0.32 
				& 0.14$^\dagger$ 
				& 0.31 
				& 0.25
				& 0.35 
				& 0.31 
				& 0.36 
				& 0.35
				\\
				
				ROUGE       & 0.34    
				& 0.22
				& 0.33  
				& 0.26
				& 0.36    
				& 0.40 
				& 0.32  
				& 0.43      
				\\
				
				METEOR      & 0.37 
				& 0.33 
				& 0.33 
				& 0.27
				& 0.37
				& 0.48 
				& 0.34   
				& 0.49      
				\\
				
				BERTScore   & 0.38 
				& 0.31  
				& 0.37
				& 0.39
				& 0.40 
				& 0.49 
				& 0.41 
				& 0.42       \\
				
				ADEM        & 0.28
				& 0.28 
				& 0.42  
				& 0.45
				& 0.26
				& 0.24
				& 0.25     
				& 0.28     \\
				
				RUBER       & 0.18$^\dagger$ 
				& 0.15$^\dagger$
				& 0.36 
				& 0.30
				& 0.33
				& 0.34 
				& 0.38   
				& 0.35       \\
				
				RoBERTa-eval& 0.68 
				& 0.71
				& 0.62  
				& 0.63
				& 0.72 
				& 0.72
				& 0.76
				& 0.77 \\
				\midrule
				\textit{Our Implementation}      \\
				ELECTRA     & 0.47
				& 0.50 
				& 0.45  
				& 0.46
				& 0.44 
				& 0.46
				& 0.52
				& 0.52                \\
				ELECTRA-DAPO& \textbf{0.76} (+29\%)
				& \textbf{0.75} (+25\%)
				& \textbf{0.77} (+32\%)    
				& \textbf{0.78} (+32\%)
				& \textbf{0.74} (+30\%)
				& \textbf{0.74} (+28\%)
				& \textbf{0.80} (+28\%)
				& \textbf{0.81} (+29\%) \\
				\bottomrule        
			\end{tabular}%
		}
		\caption{\textbf{Main results on DailyDialog and PERSONA-CHAT}: Pearson and Spearman correlation with human judgements of \textit{overall quality} on DailyDialog and PERSONA-CHAT datasets. All values that are not statistically significant (p-value $>$ 0.05) are marked by $\dagger$. Scores in bold are the best results. Following \cite{DesigningPA2020Zhao}, we divide the two datasets into train/dev/test set randomly with the ratio 0.8/0.1/0.1, and re-run all the baselines.}
	\end{table}
	
	\begin{table}
		\renewcommand\arraystretch{0.8}
		\centering
		\footnotesize
		\setlength{\tabcolsep}{3.6pt}
		\label{tb:benchmark}
		\begin{tabular}{lccccccc}
			\toprule
			\textbf{Model \textbackslash{} Dataset} & \multicolumn{3}{c}{\textbf{GRADE}} & \textbf{DSTC-6} &  \multicolumn{2}{c}{\textbf{USR}}&
			\textbf{A.V.G.}\\
			& \textbf{ConvAI2} 
			& \textbf{DailyDialog} 
			& \textbf{Empathetic} 
			&  
			& \textbf{perason-chat}  
			& \textbf{topical-chat}
			&\\
			\midrule
			BLEU & 0.00$^\dagger$ / 0.13 & 0.08$^\dagger$ / 0.18 &-0.05$^\dagger$ / 0.00$^\dagger$ &0.13 / 0.30& 0.14 / 0.09$^\dagger$ & 0.22 / 0.30 
			& 0.09 / 0.17\\
			ROUGE & 0.14 / 0.14 &0.15 / 0.15 &0.03$^\dagger$ / -0.01$^\dagger$&  0.33 / 0.33 & 0.07$^\dagger$ / 0.09$^\dagger$ & 0.28 / 0.29 
			&0.17 / 0.17\\
			METEOR &0.15 / 0.18& 0.10$^\dagger$ / 0.01$^\dagger$&0.12 / 0.06$^\dagger$&  0.31 / 0.32 & 0.25 / 0.27 & \textbf{0.34} / \textbf{0.39}
			&0.21 / 0.21\\
			BERTScore &0.23 / 0.22& 0.13 / 0.10$^\dagger$& 0.05$^\dagger$ /  0.03$^\dagger$& \textbf{0.37} / \textbf{0.34} & 0.15 / 0.12$^\dagger$ & 0.30 / 0.33
			&0.21 / 0.19\\
			ADEM &-0.06$^\dagger$ / -0.06$^\dagger$&0.06$^\dagger$ / 0.07$^\dagger$&-0.04$^\dagger$ / -0.03$^\dagger$& 0.15 / 0.12 & -0.14 / -0.09$^\dagger$ & -0.06$^\dagger$ / -0.06$^\dagger$ 
			&-0.02 / -0.01\\
			RUBER &-0.03$^\dagger$ / -0.04$^\dagger$&-0.08$^\dagger$	/ -0.09$^\dagger$ &-0.08$^\dagger$ / -0.04$^\dagger$& 0.11 / 0.09 & 0.13 / 0.19 & 0.25 / 0.26
			&0.05 / 0.06\\
			\midrule
			ELECTRA-DAPO & 0.29 / 0.30 & \textbf{0.34} / \textbf{0.33} & \textbf{0.44} / \textbf{0.43} & 0.23 / 0.24 & 0.24 / 0.21 & \textbf{0.34} / 0.30 
			&0.31 / 0.30\\
			\bottomrule
		\end{tabular}%
		\caption{\textbf{Main results on the dialogue evaluation repository in \cite{DBLP:journals/corr/abs-2106-03706}}: The metrics are formatted as \textbf{Pearson / Spearman}. The repository contains a large number of dialogue evaluation datasets. We directly apply our model and score the examples in these datasets without further training. All these datasets are QE tasks in response-level. All values that are not statistically significant (p-value $>$ 0.05) are marked by $\dagger$. 
		}
		
	\end{table}%
	
	\section{Experiments}
	\subsection{Main Results}
	Tables 2, 3, and 4 show the results on MuTual, MuTual$^{plus}$, DailyDialog, PERSONA-CHAT, and the datasets in \cite{DBLP:journals/corr/abs-2106-03706}.\footnote{DAPO has kept the state-of-the-art results in the MuTual leaderboard for three months.}
	\textcolor{black}{We also highlight the absolute improvement over the ELECTRA baseline in Tables 2 and 3.} 

	\textcolor{black}{The results in Table 2 show that ELECTRA-DAPO surpasses most of the baselines significantly. Besides, ELECTRA-DAPO gets comparable scores with previous SOTA methods (GRN-v2 and MDFN) that relies on exclusively-designed model architecture for response selection tasks, while we use a quite simple structure. The advantage in response selection task demonstrates the strong transfer learning ability of our model by modeling the fine-grain quality of dialogues in the pre-training state.}
	
	\textcolor{black}{The results in Table 3 show that ELECTRA-DAPO can be a good dialogue evaluator when the training set is relatively small. Compared with RoBerta-eval, the previous SOTA, our model is significantly better. We also see ELECTRA-DAPO gets about 30 absolute scores than the baseline of DAPO, which shows the effectiveness of incorporating dialogue-specific attributes in pre-training.}
	
	\textcolor{black}{ELECTRA-DAPO can also serves as a good zero-shot evaluator when there is no further training data. We compare ELECTRA-DAPO with several commonly-used metrics in Table 4
	, and our model correlates with the human annotation scores better. Although in some subsets ELECTRA-DAPO is not the best, it is more robust since it does not fail on any subset (all scores $>$ 0.2). The average pearson/spearman correlation score is 0.31/0.30, which is 0.1 higher than the best baseline.} 
	

	\subsection{Ablation Study}\label{sec:wo_nidf}
	As mentioned in Section \ref{sec:score_nidf}, we do \textit{specificity}-oriented automatic rescoring on positive samples in our dialogue corpus. To evaluate its contributions, we conduct an ablation study by removing it from our method and re-run the tasks. Specifically, a sample \textit{s} is scored as follows:
	$$ \textrm{score}(s)=\left\{
	\begin{aligned}
		&0                             &\ &s \in S^- \\
		&1                             &\ &s \in S^+ \\
	\end{aligned}
	\right.
	$$
	
	The results are shown in Table 5 
	. Since the weakened scoring strategy still measures \textit{coherence}, it surpasses the ELECTRA baseline. However, without \textit{token-specificity} which indicates whether a dialogue sample is \textit{Specific} and \textit{Informative}, the model becomes less powerful compared with the complete one. It also holds intuitively because the complete DAPO leverages more significant qualities of dialogues simultaneously in pre-training.  
	
	\begin{table}
		\renewcommand\arraystretch{0.8}
		\centering
		\footnotesize
		\setlength{\tabcolsep}{4.4pt}
		\label{tb:abs_nidf}
		{
			\begin{tabular}{lcccccc}
				\toprule
				\textbf{Model} 
				& \textbf{MuTual}
				& \textbf{MuTual$^{plus}$}
				
				& \multicolumn{2}{c}{\textbf{DailyDialog}}
				& \multicolumn{2}{c}{\textbf{PERSONA-CHAT}}\\
				
				\ 
				& 
				& 
				
				& \multicolumn{4}{c}{\textbf{Pearson} / \textbf{Spearman}}
				\\
				
				&\textbf{R@1} / \textbf{R@2} / \textbf{MRR} &\textbf{R@1} / \textbf{R@2} / \textbf{MRR} &\textbf{Dev} &\textbf{Test} &\textbf{Dev} &\textbf{Test}\\
				\midrule
				ELECTRA 
				
				& 0.900 / 0.979 / 0.946
				& 0.823 / 0.947 / 0.901
				
				& 0.47 / 0.50
				& 0.45 / 0.46
				& 0.44 / 0.46
				& 0.52 / 0.52 \\
				
				ELECTRA-DAPO w/o TS 
				
				& 0.903 / 0.980 / 0.947
				& 0.819 / 0.945 / 0.899
				
				& 0.66 / 0.67
				& 0.69 / 0.70
				& 0.63 / 0.65
				& 0.66 / 0.72 \\
				
				ELECTRA-DAPO (3-NIDF)
				
				& \textbf{0.916} / \textbf{0.988} / \textbf{0.956}
				& \textbf{0.836} / 0.955 / \textbf{0.910}
				
				& \textbf{0.76} / \textbf{0.75}
				& \textbf{0.77} / \textbf{0.78}
				& \textbf{0.74} / \textbf{0.74}
				& \textbf{0.80} / \textbf{0.81} \\
				
				ELECTRA-DAPO (2-NIDF) 
				
				& 0.907 / 0.981 / 0.950
				& 0.827 / \textbf{0.958} / 0.902
				
				& 0.75 / 0.73
				& 0.65 / 0.69
				& 0.65 / 0.65
				& 0.71 / 0.73 \\
				
				ELECTRA-DAPO (1-NIDF) 
				
				& 0.904 / 0.980 / 0.949
				& 0.819 / 0.940 / 0.904
				
				& 0.73 / 0.65
				& 0.55 / 0.60
				& 0.63 / 0.60
				& 0.65 / 0.69 \\
				
				\bottomrule       
			\end{tabular}
		}
		\caption{\textbf{Effect of different rescoring module:} We re-run the 	experiments by replacing the automatic rescoring module with other variants, w/o \textit{TS} refers to without \textit{token-specificity}.}
		
	\end{table}
	
	\begin{figure}
		\centering
		\includegraphics[width=0.6\textwidth]{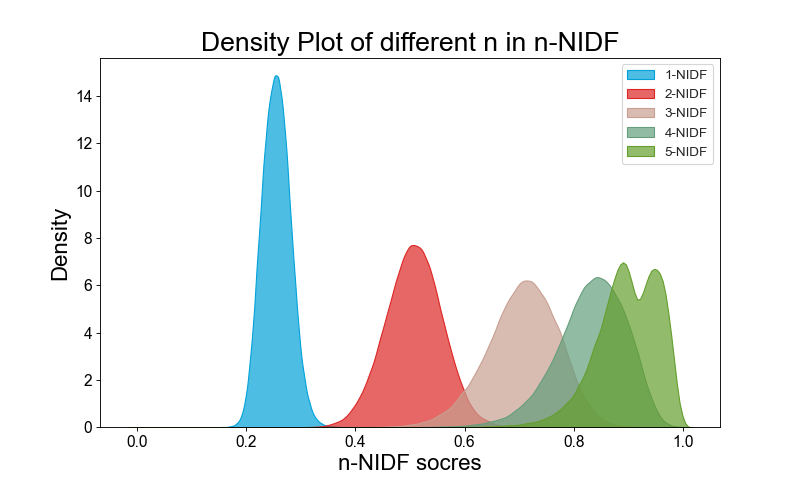}
		\caption{\textcolor{black}{Distribution of \textit{N}-NIDF scores.}}
		\label{fig:dis_NIDF}
	\end{figure}
	
	\subsection{The influence of \textit{N} in \textit{N}-NIDF}
	Results in Section \ref{sec:wo_nidf} show the importance of \textit{specificity}-oriented automatic rescoring in DAPO, thus it is reasonable to investigate the influence of \textit{N} when calculating \textit{N}-NIDF. We re-run the experiments by using 2-NIDF and 1-NIDF as the \textit{token-specificity} coefficient. The results are also shown in Table 5 
	, indicating that DAPO with larger $N$ in $N$-NIDF tends to have a better performance. We further explore the distribution of \textit{N}-NIDF scores in our corpus. \textcolor{black}{Figure \ref{fig:dis_NIDF} shows clearly that the distribution of the 1-NIDF score is more concentrated than those for 2 and 3-NIDF scores, which makes it harder for the model to learn how to distinguish different dialogue samples. This may be a potential explanation for the above observation. It is also found that 1- to 4-NIDF scores generally follow a normal distribution. To some extent, we believe this reflects the general pattern of human dialogues. The distribution of the 4- and 5-NIDF are similar to 3-NIDF, except that the distribution of 5-NIDF has two peaks. We try $N$=3,4,5 on the zero-shot QE tasks, and the results are shown in Table 6} 
	. We can clearly see that for $N\ge3$, the performance remains virtually unchanged, so using $N=3$ is quite suitable.

	\begin{table}[htbp]
		\renewcommand\arraystretch{0.8}
		\footnotesize
		\centering
		\label{tb:more_comp}%
		\begin{tabular}{ccc}
			\toprule
			\textbf{Models \textbackslash{} Correlation} & \textbf{Pearson} & \textbf{Spearman} \\
			\midrule
			ELECTRA + DAPO (3-NIDF) & 0.314 & 0.302 \\
			ELECTRA + DAPO (4-NIDF) & 0.313 & 0.307 \\
			ELECTRA + DAPO (5-NIDF) & 0.307 & 0.304 \\
			\bottomrule
		\end{tabular}%
		\caption{\textbf{More effects of different $N$ in $N$-NIDF}: We try 3,4,5 NIDF and get average Pearson and Spearman correleation values for datasets in the repository \cite{DBLP:journals/corr/abs-2106-03706}}
	\end{table}%
	
	\begin{table}
		\renewcommand\arraystretch{0.8}
		\centering
		\footnotesize
		\setlength{\tabcolsep}{3.5pt}
		\label{tb:ubuntu_pretrain}
		{
			\begin{tabular}{lcccccc}
				\toprule
				\textbf{Model} 
				& \textbf{MuTual}
				& \textbf{MuTual$^{plus}$}
				
				& \multicolumn{2}{c}{\textbf{DailyDialog}}
				& \multicolumn{2}{c}{\textbf{PERSONA-CHAT}}\\
				
				\ 
				& 
				& 
				
				& \multicolumn{4}{c}{\textbf{Pearson} / \textbf{Spearman}}
				\\
				
				&\textbf{R@1} / \textbf{R@2} / \textbf{MRR} &\textbf{R@1} / \textbf{R@2} / \textbf{MRR} &\textbf{Dev} &\textbf{Test} &\textbf{Dev} &\textbf{Test}\\
				\midrule
				
				ELECTRA-DAPO (UI+UR+UO)
				
				& \textbf{0.916} / \textbf{0.988} / \textbf{0.956}
				& \textbf{0.836} / \textbf{0.955} / \textbf{0.910}
				
				& \textbf{0.76} / \textbf{0.75}
				& \textbf{0.77} / \textbf{0.78}
				& \textbf{0.74} / \textbf{0.74}
				& \textbf{0.80} / \textbf{0.81} \\

				ELECTRA-DAPO (Ubuntu)
				
				& 0.274 / 0.519 / 0.539
				& 0.284 / 0.537 / 0.547
				
				& 0.09 / 0.11
				& 0.25 / 0.21
				& 0.20 / 0.16
				& 0.11 / 0.11 \\
				
				ELECTRA-DAPO (only UI)
				
				& 0.898 / 0.977 / 0.945
				& 0.819 / 0.947 / 0.900
				
				& 0.71 / 0.68
				& 0.60 / 0.60
				& 0.61 / 0.63
				& 0.66 / 0.67 \\
				
				ELECTRA-DAPO (only UR)
				
				& 0.900 / 0.972 / 0.945
				& 0.821 / 0.944 / 0.900
				
				& 0.73 / 0.70
				& 0.72 / 0.73
				& 0.67 / 0.67
				& 0.75 / 0.77 \\
				
				ELECTRA-DAPO (only UO)
				
				& 0.892 / 0.970 / 0.940
				& 0.808 / 0.947 / 0.894
				
				& 0.60 / 0.58
				& 0.45 / 0.46
				& 0.44 / 0.45
				& 0.48 / 0.51 \\
				\bottomrule       
			\end{tabular}
		}
		\caption{\textbf{Effect of sample construction module and data source:} We re-run the experiments by using only one of the methods in \textit{coherence}-oriented context corruption to construct the training set, and also try Ubuntu as the data source which contains much more noise.}
		
	\end{table}
	


	\subsection{Construction of pre-training corpus} 
	Since the corpus of pre-training is newly generated, we further explore how different constructing methods influence the performance of our models. 
	
	We first re-construct another pre-training corpus with Ubuntu \cite{lowe2015ubuntu}, which is noisy and not finely proofread. We also modify the way of generating negative samples by using only one of UO, UI and UR to do \textit{coherence}-oriented context corruption and construct 3 more pre-training corpora. The proportion of positive and negative samples are kept the same as before, i.e., $|S^+|$ : $|S^-|$ = 1 : 3. Experiments are done on these four new pre-training corpora and the results are shown in Table 7 
	. Clearly, DAPO with all three kinds of negative samples boosts our models the most, while using only one of the generating methods hurts the model performance more or less even with the same sized corpus. Accordingly, we argue that training models to tackle different kinds of negative samples in a single pre-training task makes them more robust. Thus the models can learn to address more complex dialogues through DAPO. We also see that pre-training with the noisy data, Ubuntu, degrades the performance extremely. This can be largely attributed to the low \textit{Readability} and \textit{coherence} of Ubuntu dataset. Consequently, the model fails on the pre-training task (only 0.009 and 0.011 for the Pearson and Spearman correlation metrics) and, in turn, on the downstream tasks. 
	
	\begin{table}
		\renewcommand\arraystretch{0.8}
		\centering
		\footnotesize
		\setlength{\tabcolsep}{12pt}
		\label{tb:diff_obj}
		{
			\begin{tabular}{lcccc}
				\toprule
				\textbf{Model} 
				& \textbf{MuTual}
				& \textbf{MuTual$^{plus}$}
				
				& \textbf{DailyDialog}
				& \textbf{PERSONA-CHAT}
				\\
				
				&\textbf{R@1} / \textbf{R@2} / \textbf{MRR} &\textbf{R@1} / \textbf{R@2} / \textbf{MRR} &\textbf{Pearson}/\textbf{Spearman} &\textbf{Pearson}/\textbf{Spearman}\\
				\midrule
				
				ELECTRA
				
				& 0.900 / 0.979 / 0.946
				& 0.823 / 0.947 / 0.901
				
				& 0.45 / 0.46
				& 0.52 / 0.52
				\\
				
				+MLM
				
				& 0.847 / 0.955 / 0.915
				& 0.737 / 0.916 / 0.853
				
				& 0.31 / 0.32
				& 0.51 / 0.52
				\\
				
				+NSP
				
				& 0.903 / 0.979 / 0.949
				& 0.823 / 0.941 / 0.900
				
				& 0.69 / 0.70
				& 0.70 / 0.71
				\\
				
				+MLM+NSP
				
				& 0.891 / 0.964 / 0.939
				& 0.818 / 0.942 / 0.898
				
				& 0.67 / 0.71
				& 0.68 / 0.70
				\\
				\midrule
				+DAPO
				
				& \textbf{0.916} / \textbf{0.988} / \textbf{0.956}
				& \textbf{0.836} / \textbf{0.955} / \textbf{0.910}
				
				& \textbf{0.77} / \textbf{0.78}
				& \textbf{0.80} / \textbf{0.81}
				\\
				
				\bottomrule       
			\end{tabular}
		}
		\caption{
			\textbf{Effects of different pre-training objectives:} We pre-train our model on the same data source with different training objectives, and re-run the downstream tasks, the results on DailyDialog and PERASONA-CHAT are test sets.}
	\end{table}
	
	\subsection{Comparison with Different Pre-training Objectives}
	We compare DAPO with representative \textit{coherence}-oriented objectives mentioned in Section \ref{sec:baseline}. Table 8 
	shows the results. According to the results, ELECTRA-MLM, although dialogue-adaptive pre-trained, has much worse performance than ELECTRA on almost all these open-domain dialogue tasks, which indicates that MLM would not be a suitable pre-training objective for dialogue-based texts. The performance of ELECTRA-NSP is between ELECTRA and ELECTRA-DAPO, which indicates that NSP is a feasible pre-training objective for dialogues. For ELECTRA-MLM+NSP, since it combines both proper and improper objectives, it is reasonable that it has performance between ELECTRA-MLM and ELECTRA-NSP.
	
	In addition, to verify if this kind of difference is caused by an insufficiency of pre-training, so we evaluate the models on the dev set of its pre-training corpus. The Pearson and Spearman correlation of ELECTRA-DAPO is 0.810 and 0.690 respectively; the accuracy of ELECTRA-NSP and ELECTRA-MLM+NSP are 94.7\% and 92.8\% respectively; and the perplexity of ELECTRA-MLM is 4.96. These values show that models are all fully pre-trained towards the given objectives. The findings above gave another evidence that the performance of models mainly depends on whether its pre-training objective is suitable for an open-domain dialogue corpus.

	\section{Conclusion}
	In this work, we propose dialogue-adaptive pre-training objectives from quality estimation to capture exclusive attributes for open-domain dialogues. Through \textit{coherence}-oriented context corruption and \textit{token-specificity} rescoring steps, we fuse quality estimation factors into pre-training, which enables models to learn more compatible language representations for open-domain dialogue tasks. Experiments on widely-used open-domain dialogue datasets show our superiority over baseline methods. 
	Beyond the common practice that merely uses general training objectives in domain-adaptive pre-training, our work further incorporates the specific features of the in-domain texts into pre-training tasks, empowering the PrLMs by modeling the diverse aspects of dialogue-related characteristics.

	\bibliography{reference}
	
\end{document}